%% file: main.tex
\documentclass[10pt,twocolumn,letterpaper]{article}

\usepackage[pagenumbers]{iccv} 


\definecolor{iccvblue}{rgb}{0.21,0.49,0.74}
\usepackage[pagebackref,breaklinks,colorlinks,allcolors=iccvblue]{hyperref}

\def\paperID{7464} 
\def\confName{ICCV}
\def\confYear{2025}

\title{Is Pre-training Applicable to the Decoder for Dense Prediction?}

\author{Chao Ning\\
The University of tokyo\\
\and
Wanshui Gan\\
The University of tokyo\\
\and
Weihao Xuan\\
The University of tokyo\\
\and
Naoto Yokoya\\
The University of tokyo\\
}

\usepackage{multirow}

\begin{document}
\maketitle 

\begin{abstract}
Encoder-decoder networks are commonly used model architectures for dense prediction tasks, where the encoder typically employs a model pre-trained on upstream tasks, while the decoder is often either randomly initialized or pre-trained on other tasks. In this paper, we introduce $\times$Net, a novel framework that leverages a model pre-trained on upstream tasks as the decoder, fostering a ``pre-trained encoder $\times$ pre-trained decoder'' collaboration within the encoder-decoder network. $\times$Net effectively address the challenges associated with using pre-trained models in the decoding, applying the learned representations to enhance the decoding process. This enables the model to achieve more precise and high-quality dense predictions.
By simply coupling the pre-trained encoder and pre-trained decoder, $\times$Net distinguishes itself as a highly promising approach. Remarkably, it achieves this without relying on decoding-specific structures or task-specific algorithms. Despite its streamlined design, $\times$Net outperforms advanced methods in tasks such as monocular depth estimation and semantic segmentation, achieving state-of-the-art performance particularly in monocular depth estimation.
\end{abstract}


\section{Introduction}
\label{sec:intro}
Since 2015, Jonathan et al.~\cite{long2015fully} have reinterpreted classification networks as fully convolutional architectures, fine-tuning these models based on their pre-learned representations. Pre-trained models excel at extracting features across multiple scales, from fine to coarse, effectively capturing both local and global information from images.

In the years that followed, numerous studies have explored improved network structures to decode these visual features, establishing end-to-end networks that provide effective solutions for various dense prediction tasks such as monocular depth estimation~\cite{godard2017unsupervised,lee2019big,bhat2021adabins,shao2023nddepth}, semantic segmentation~\cite{lin2017refinenet,zhao2017pyramid,xiao2018unified}, and optical flow estimation~\cite{mayer2016large, weinzaepfel2023croco, xu2023unifying}. In these tasks, pre-trained models act as encoders to extract features, which are progressively refined from coarse visual representations to pixel-level predictions through randomly initialized decoding techniques.

Recently, networks trained with larger datasets have emerged as promising solutions for a variety of computer vision tasks, including image classification~\cite{sun2017revisiting}, self-supervised learning~\cite{oquab2023dinov2}, semantic segmentation~\cite{kirillov2023segment}, and depth prediction~\cite{yang2024depth,hu2024metric3d}.
The expansion of training data has resulted in significant enhancements in network performance, enabling classical network structures to surpass even more advanced architectures. For example, in the field of monocular depth estimation, Depth Anything~\cite{yang2024depth} employs the 2021 architecture of DPT~\cite{ranftl2021vision}, trained with larger datasets, and is able to surpass the state-of-the-art architectures of 2023, such as NDDepth~\cite{shao2023nddepth}.

\begin{figure}
\centering
\includegraphics[width=.45\textwidth]{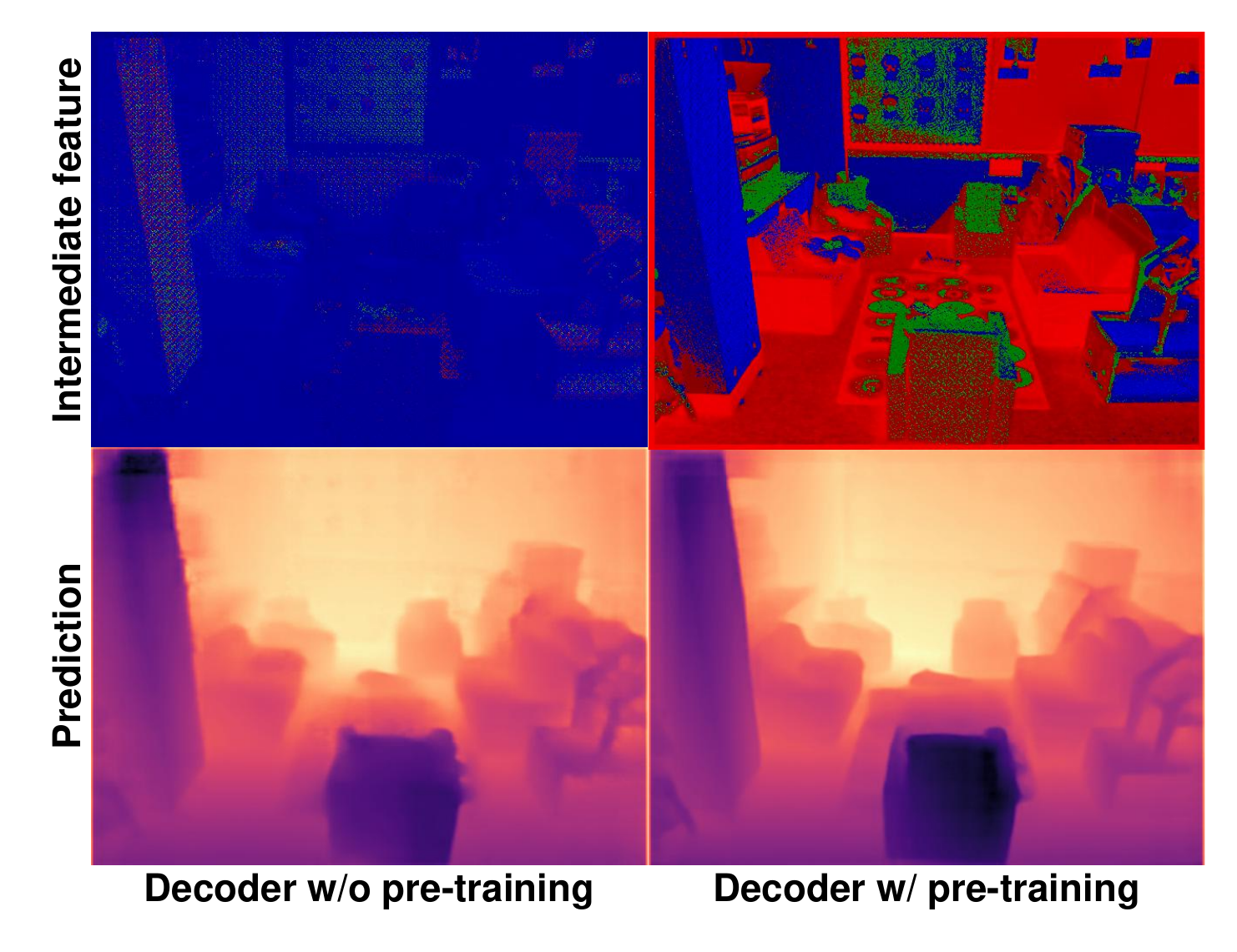}
\caption{
Comparison of pre-trained and non-pre-trained decoders.
When paired with a consistent encoder and architecture, a pre-trained decoder enhances the semantic richness of encoded feature maps and refines the final predictions.
}
\label{fig:start}
\end{figure}

Reflecting on encoder-decoder networks for dense prediction, the encoder is typically a pre-trained model, while the decoder is usually initialized randomly. This raises an important question: given that increased training data can enhance network performance, why is the decoder not pre-trained as well?

To address this, we investigate why encoders benefit from pre-training to improve performance, whereas decoders typically do not.

To use a pre-trained model as a decoder, two important challenges must be addressed. First, (1) the decoder's structure inherently differs from that of the pre-trained model, preventing it from loading pre-trained parameters. Second, (2) as the decoding component, the decoder must derive dense predictions from features, which differs from the encoder's goal of extracting features from images. Therefore, the parameters pre-trained on upstream tasks may not necessarily be suitable for decoding.

In this paper, we introduce $\times$Net,
which facilitates a ``pre-trained encoder $\times$ pre-trained decoder'' collaboration.
It addresses the challenges of using a pre-trained model as a decoder, enabling pre-learned knowledge to be utilized in the decoder of a dense prediction network, thereby achieving significant improvements.
As shown in Figure~\ref{fig:start}, $\times$Net enriches the network's intermediate feature maps with semantic information. This semantic information, which is not supervised during dense prediction tasks, can only be acquired through pre-training and aids in achieving higher-quality dense predictions. When magnified, it is clearly observable that the pre-trained decoder exhibits sharper edges and finer details.



In summary, our core contributions are as follows:

\begin{itemize}
\item To the best of our knowledge, we are the first to propose applying a pre-trained model as a decoder in dense prediction network and to address the challenges associated with using pre-trained models in the decoding.
\item We explored why a pretrained decoder is necessary for dense prediction, emphasizing its ability to provide semantic information crucial for achieving high-quality outputs. 
\item Extensive experiments demonstrate that using a pre-trained decoder, without additional complex task-specific designs, can achieve advanced performance.
\end{itemize}

\begin{figure*}
\centering
\includegraphics[width=1.\textwidth]{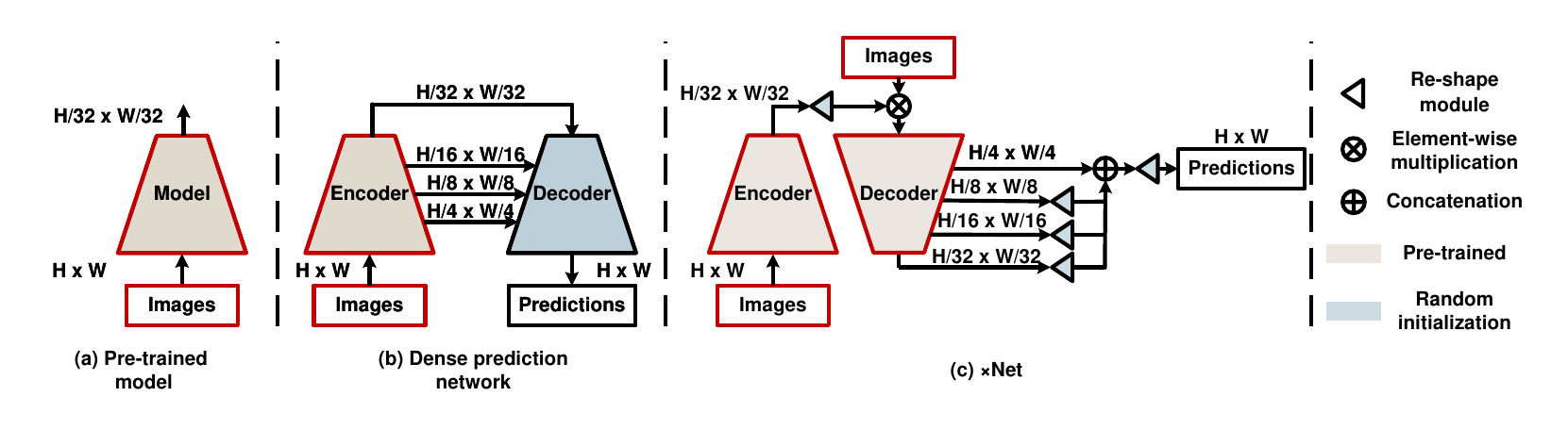}
\caption{
Overview of the image classification model (left), encoder-decoder network (middle), and our $\times$Net (right).
The pre-trained model represents the commonly used hierarchical structure.
}
\label{fig:Overview}
\end{figure*}

\section{Related work}

\subsection{Model pre-training}
Model pre-training facilitates the learning of visual representations, improving the performance of vision models in downstream tasks. Supervised training for image classification is a classical and widely utilized pre-training method, with classification accuracy serving as an evaluative metric for model performance. ImageNet-1k~\cite{russakovsky2015imagenet} training, which includes 1.28 million training images across one thousand categories, is a standard regimen. To enable networks to learn richer visual representations, image classification with more categories and a greater number of images is employed in pre-training. To effectively enhance the quality of pre-training, methods such as BEiT~\cite{beit}, which utilize masked image modeling, CLIP~\cite{radford2021learning}, which employs image-language data pairs for classification training, and DINO~\cite{caron2021emerging, oquab2023dinov2}, which focus on self-supervised learning for vision transformers, leverage extensive datasets to make the models more robust.
Previously, these pre-trained networks could only serve as encoders in dense prediction networks for feature extraction. In this paper, we are the first to utilize these networks as decoders to decode encoded features.


\subsection{Encoder-decoder networks for dense prediction}
The pioneering work~\cite{long2015fully} first employed a decode head to fine-tune networks that achieved success in image benchmarks, specifically for pixel-level semantic segmentation. This innovative approach inspired the advancement of numerous dense prediction tasks, which now focus on the efficient decoding of features extracted by pre-trained encoders. To effectively capture multi-scale information, Feature Pyramid Networks (FPN)~\cite{kirillov2019panoptic, lin2017feature} incorporate encoded features at various resolutions into the decoding process. This straightforward yet effective concept has become a fundamental architecture widely adopted in dense prediction networks. Based on the FPN architecture, some works enhance network performance by improving network structures. These improvements include plug-in modules~\cite{zhao2017pyramid, chen2018encoder, huang2019ccnet,wang2018non}, decoding architecture~\cite{ranftl2021vision} adapted to columnar Vision Transformer (ViT) encoders, and decoders utilizing transformer architectures~\cite{cheng2021per, cheng2022masked, yuan2022neural, li2024binsformer}. On the other hand, the dense prediction performance is improved through specialized algorithms tailored to specific tasks. For instance, customized context~\cite{context1,context2,context3} in semantic segmentation, clustering algorithms~\cite{cluster1, cluster2} in instance segmentation, bins classification~\cite{bhat2021adabins,shao2024iebins,bhat2022localbins}, normal distance assistance~\cite{shao2023nddepth}, and ground embedding~\cite{yang2023gedepth} in monocular depth estimation. Our experiments demonstrate that introducing a pre-trained decoder can surpass advanced decoding structures and specialized algorithms.

\section{From pre-trained model to decoder}
In this section, we investigate how to effectively integrate pre-trained model into the decoder, addressing two major challenges.
First, common dense prediction network decoders cannot utilize pre-trained parameters.
Second, even if pre-trained parameters can be loaded into the decoder, ensuring effective use of the pre-learned representations during the decoding process remains a challenge.
We conduct a theoretical analysis of these two challenges, propose solutions, and finally summarize our method, introducing $\times$Net, a dense prediction network that efficiently leverages a pre-trained decoder.
We select the widely utilized hierarchical model pre-trained on image classification tasks to represent the pre-trained models.
Additionally, we employ an encoder-decoder network with hierarchically connected pyramidal structures to represent common dense prediction networks.

\subsection{Why pre-training is not available for decoder?}

\textbf{Reversed structure.}
As illustrated in Figure~\ref{fig:Overview},
the primary reason pre-training is not applied to decoders is that the structure of decoders is reversed compared to that of pre-trained models.
Pre-trained models typically reduce the spatial size of features progressively to extract image categories, whereas dense prediction requires reconstructing high-resolution pixel-level predictions from low-resolution feature maps.
The resolutions of feature maps in traditional decoders and pre-trained models can be described as follows:
\begin{equation}
\left\{
\begin{array}{ll}
F^{\frac{1}{32}} \xrightarrow{\mathcal{D}_1} F^{\frac{1}{16}} \xrightarrow{\mathcal{D}_2} F^{\frac{1}{8}}\xrightarrow{\mathcal{D}_3}
F^{\frac{1}{4}}
\xrightarrow{\mathcal{D}_4, \text{head}} F^{1}
\\
F^{1} \xrightarrow{\text{stem}, \mathcal{P}_1} F^{\frac{1}{4}} \xrightarrow{\mathcal{P}_2} F^{\frac{1}{8}}\xrightarrow{\mathcal{P}_3}
F^{\frac{1}{16}}
\xrightarrow{\mathcal{P}_4} F^{\frac{1}{32}}
\end{array}
\right .
.
\label{eq:1}
\end{equation}
Here, $\mathcal{D}$ and $\mathcal{P}$ denote the stages of the decoder and the pre-trained model, respectively, with the subscript indicating their sequential order within the model.
$F$ represents the feature map, and the superscript denotes the downsampling factor of the width and height.
The reversed decoding structure is a prerequisite for utilizing pre-trained parameters.
Obviously, employing a pre-trained decoder requires input features at the original resolution and output pixel-level predictions.

\textbf{Reversed decoder input.}
As described in Equation~\ref{eq:1}, the input to the decoder is $F^{\frac{1}{32}}$, which must be reshaped to an appropriate resolution, such as $F^{1}$, to be compatible with the pre-trained decoder.
We propose a Re-Shape (RS) module to directly re-shape the encoded feature maps to high resolution for reversed decoder structure.
We propose a simple Re-Shape (RS) module that directly reshapes the encoder output to a size suitable for input into the reversed decoder structure. Specifically, the RS module consists of a normalization layer and a two-layer MLP, with upsampling achieved through a PixelShuffle~\cite{shi2016real} module positioned of the MLP.

\textbf{Reversed decoder prediction.}
According to Equation~\ref{eq:1}, the traditional decoder is required to produce pixel-level predictions ($F^1$). Achieving pixel-level predictions directly from the output of a pre-trained model ($F^{\frac{1}{32}}$) is challenging. Therefore, we propose the Post Feature Pyramid (PFP), which reshapes the feature maps from each of the four stages of the pre-trained decoder to the same resolution using the RS module. These reshaped features are then concatenated along the channel dimension and subsequently fused through another RS module.

With the aforementioned adjustments, the pre-trained model can be used as a decoder, enabling pixel-level predictions in dense prediction tasks.

\subsection{How to fine-tune pre-trained decoder?}
\label{sec:3.2}
\textbf{Optimization challenges.}
To effectively fine-tune a pre-trained decoder and fully leverage its learned representations during decoding, it is important to mitigate the optimization challenges. As illustrated in Figure~\ref{fig:Overview}, decoders in common dense networks are required to process features from multiple stages of the encoder's output. In contrast, pre-training typically involves extracting features and recognizing objects from image inputs. Thus, to facilitate optimization, decoding should be aligned in a manner that complements the pre-learned representations.

\textbf{Optimization of decoder stages.}
In dense prediction networks, the pyramidal connections between the encoder and decoder can introduce optimization difficulties for the pre-trained decoder.
The traditional feature pyramid structure can be defined as follows:
\begin{equation}
F^{\frac{1}{k}} = \text{D}_i(E_i^{\frac{1}{k}} + \text{D}_{i-1}(F^{\frac{1}{2k}})), 2 \leq i \leq 4.
\end{equation}
Here $\text{D}_i$ is the $i$th decoder stage,
$F$ denotes the feature map of the decoder,
$E_i$ is encoded feature map for $i$th decoder stage and $\frac{1}{k}$ denotes a down-sampling factor of the width and height.
Consider \(E_i\) input into pre-trained $\text{D}_i$ as follows:
\begin{equation}
E_i = \phi_i(X_i),
\label{eq:op1}
\end{equation}
where \(\phi_i\) represents the variation from the pre-learned representation \(X_i\).
For each \(\text{D}_i\), there is an variation as follows:
\begin{equation}
\Delta_i = 
\| \text{D}_i(E_i) - \text{D}_i(X_i) \|^2,
\label{eq:op2}
\end{equation}
For \(\text{D}_2, \text{D}_3\), and  \(\text{D}_4\), optimization can be approximately defined as:
\begin{equation}
\min_{\text{D}_i} \, \| \text{D}_i(E_i) - \text{D}_i(X_i) \|^2 + \| \text{D}_{i}(\Delta_{i-1}) \|^2.
\end{equation}
Here the input $E_i$ of FP introduces an additional term to the optimization objective ($\| \text{D}_i(E_i) - \text{D}_i(X_i) \|^2$). Hence, we remove the FP structure for eliminating the optimization of $\| \text{D}_i(E_i) - \text{D}_i(X_i) \|^2$.

\textbf{Optimization of decoder input.}
According to Equation~\ref{eq:op1},
the variation of decoder input can be defined as:
\begin{equation}
\phi(I) = \hat{\text{E}}(I),
\end{equation}
where $I$ denotes the input image, and $\hat{\text{E}}$ is the encoder.
Then 
the optimization target can be approximately described as follows:
\begin{equation}
\left\{
\begin{array}{ll}
\min_{\text{D}_1} \, \| \text{D}_{1}(\hat{\text{E}}(I)) - \text{D}_{1}(I) \|^2 \\
\min_{\text{D}_i} \, \| \text{D}_{i}(\Delta_{i-1}) \|^2, 2 \leq i \leq 4
\end{array}
\right.
.
\end{equation}
The key to reducing optimization difficulty lies in ensuring that $D_i$ processes encoded features in a manner consistent with image processing. 
For instance, if at a certain stage the decoding of encoded features is equivalent to processing the image, subsequent stages can align with the pre-trained representations without further optimization.
However, retrieving image information from encoded features is challenging, as the encoder's output is abstract and not a simple transformation of the image. Therefore, we mix the input to the pre-trained decoder with the original image, integrating it into the features through element-wise multiplication. 
This allows $\hat{\text{E}}(I)$ to be considered a transformation of $I$,
and optimization can progress along the trajectory of this transformation.

With these designs, the decoding process more closely mirrors the image processing of pre-training, reducing the difficulty of optimization and facilitating the transfer of pre-learned representations to the decoding.

\subsection{Dense prediction using pre-trained decoder}
\begin{figure}
    \centering
\includegraphics[width=0.45\textwidth]{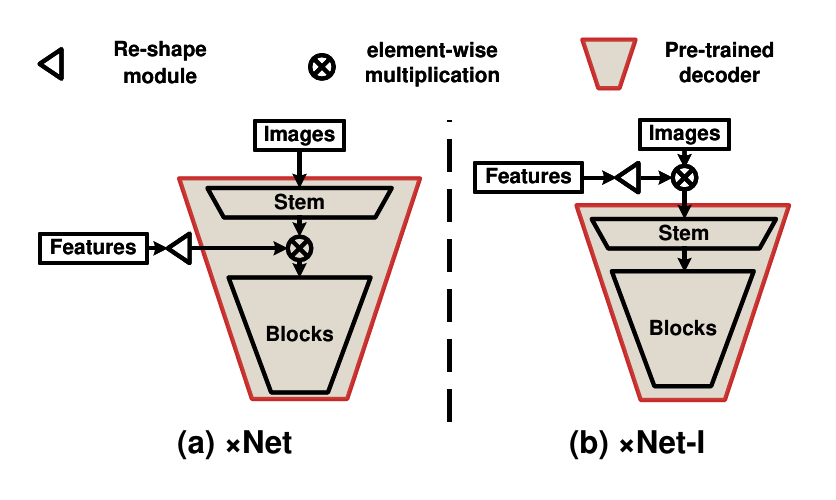}
    \caption{The difference between $\times$Net and $\times$Net-I.}
    \label{fig:mix}
\end{figure}

In this subsection, we summarize the solutions to the challenges faced by pre-trained decoders in dense prediction and introduce $\times$Net, a network capable of performing dense prediction with a ``pre-trained Encoder $\times$ pre-trained Decoder'' architecture.
As illustrated in Figure~\ref{fig:Overview}, compared to conventional dense networks, $\times$Net features a reversed decoder and employs a PFP to achieve pixel-level predictions. By removing the pyramidal connections between the encoder and decoder and integrating image data into the encoded features, the pre-trained representations are more effectively incorporated into the decoding process.
The red-highlighted parts in Figure~\ref{fig:Overview} clearly show that $\times$Net's encoder-decoder structure and workflow closely resemble that of the pre-trained model.
For the method of mixing encoded features into image processing, we propose two approaches, as illustrated in Figure~\ref{fig:mix}.
The first approach, used in $\times$Net, mixes the encoded features with the output of the stem. The second approach, applied in $\times$Net-I, mixes the encoded features directly with the RGB image. The mixing method of $\times$Net-I provides an intuitive visualization analysis strategy; however, since it compresses the encoded features to a smaller size (approximately $\frac{1}{4}$ to $\frac{1}{2}$) of the pixel count of the stem features, hence its performance is slightly inferior to that of $\times$Net.

\begin{figure*}
\centering
\includegraphics[width=\textwidth]{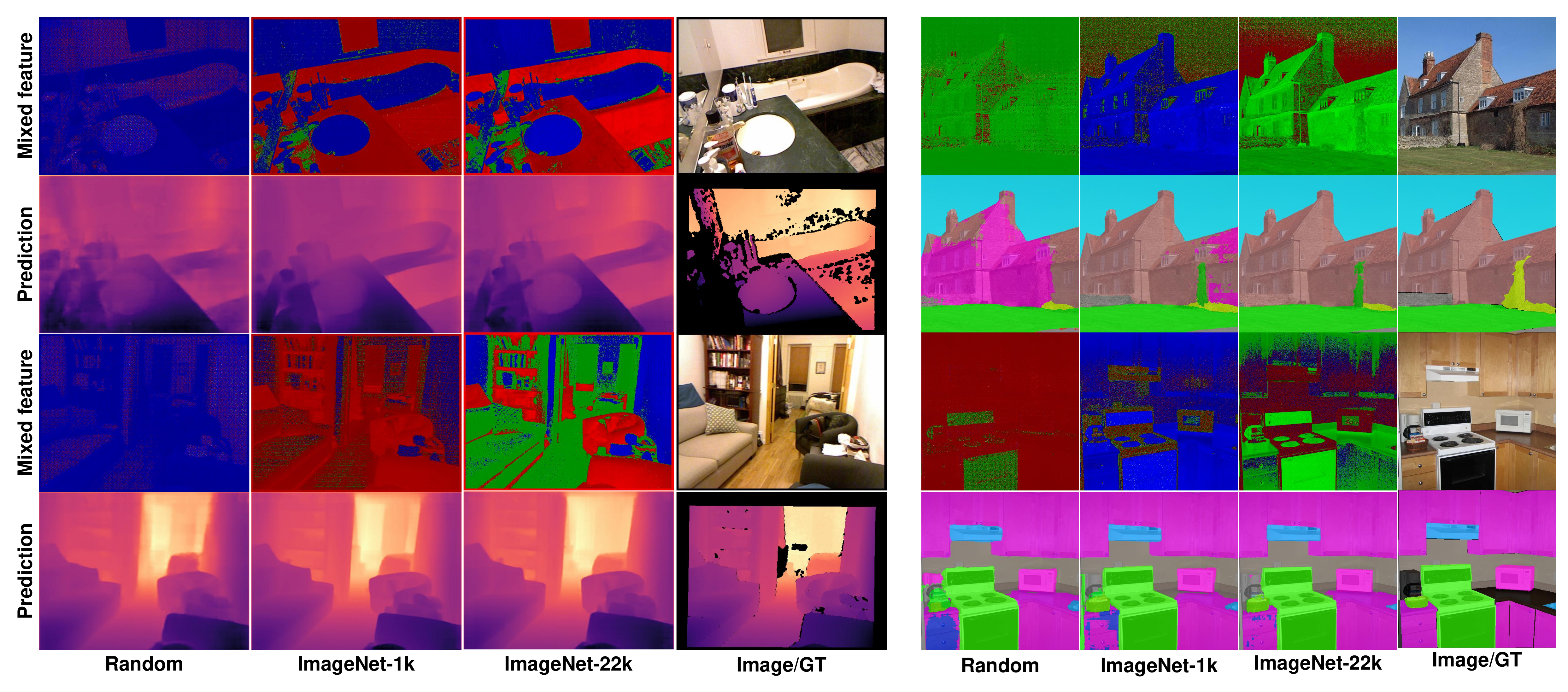}
\caption{
Qualitative comparisons of decoder pre-training on monocular depth estimation (left) and semantic segmentation (right).
For mixed feature visualization, extract the maximum value from the RGB components, setting the remaining values to zero.
}
\label{fig:nyu&seg}
\end{figure*}


\section{Experiment settings}
In this section, we present the dense prediction tasks discussed in this paper, the datasets employed, and the experimental setup.

\subsection{Dense prediction tasks}
Similar to previous works~\cite{ranftl2021vision,ji2023ddp} on dense prediction, we select two popular dense prediction tasks: monocular depth estimation (MDE) and semantic segmentation.

\textbf{MDE.} MDE is a pixel-level regression task that involves predicting 3D depth information from a single 2D image. 
The widely used approach for this task employs a pre-trained encoder to extract features from the image, followed by a decoder module, which is randomly initialized, to decode depth information from the extracted features.
Recent methods achieve improved depth prediction by designing advanced model architectures, while others use camera parameters to derive geometric information from images to assist in more accurate depth regression.

\textbf{Semantic segmentation.} 
Semantic segmentation is a pixel-level classification task that requires predicting the precise category for each pixel.
Due to the potentially large differences in target sizes, typical semantic segmentation methods utilize a multi-branch decoding structure to capture richer contextual information, enabling accurate classification for each pixel.

\subsection{Datasets and settings}
To comprehensively evaluate our approach, we conducted experiments on four monocular depth estimation (MDE) datasets and two semantic segmentation datasets.

\textbf{MDE settings.} We conducted training on three classic monocular depth estimation datasets: NYU-Depth-V2~\cite{silberman2012indoor}, KITTI~\cite{geiger2013vision}, and DDAD~\cite{guizilini20203d}, using the same training and validation splits as the comparison methods. Additionally, following previous works~\cite{bhat2021adabins, yuan2022newcrfs, shao2023nddepth}, the model trained on the indoor dataset NYU-Depth-V2 is subjected to zero-shot validation on another indoor dataset, SUN RGB-D~\cite{song2015sun}.
We follow the data preprocessing described in \cite{li2024binsformer} for the NYU-Depth-V2 and KITTI datasets, and follow the procedures outlined in~\cite{yang2023gedepth} for the DDAD dataset.
In NYU-Depth-V2, KITTI, and DDAD datasets, the maximum depth is capped at 10m, 80m, and 200m, respectively.

\textbf{Semantic segmentation settings.}
We performed semantic segmentation experiments on the ADE20K~\cite{zhou2017scene} dataset, which contains 150 categories, and the Cityscapes~\cite{cordts2016cityscapes} dataset, comprising 19 categories.
Following previous work~\cite{liu2021swin},
we use the mmsegmentation toolbox~\cite{mmseg2020} for standard training.
The crop size for ADE20K is 512$\times$512, and for Cityscapes, it is 512$\times$1024.

\section{Role of pre-trained decoders}
In this section, we first conduct ablation experiments on $\times$Net to evaluate the impact of each design component on the model's performance. We then use $\times$Net-I to analyze how the pre-trained decoder contributes to performance improvements in dense prediction tasks.
All encoders are ImageNet-22k pre-trained ConvNeXt-T~\cite{liu2022convnet}, and all decoders are ConvNeXt-T based stuructures.

\begin{table}
\resizebox{\columnwidth}{!}{
\begin{tabular}{l|cc|cc}
\midrule
\multirow{2}{*}{Variant} & \multicolumn{2}{|c|}{NYU MDE}&\multicolumn{2}{|c}{ADE20k Seg.} \\
 & $\delta<1.25 \uparrow$ & Abs Rel $\downarrow$ & mAcc$\uparrow$ & mIoU $\uparrow$\\
\midrule
\midrule
FPN & 0.889 & 0.106 & 52.92 & 42.82 \\
+Rev. & 0.896 & 0.104 & 52.42 & 42.31 \\
+22k & 0.909 & 0.099 & 55.32 & 44.71 \\
+PFP & 0.910 & 0.098 & 55.47 & 44.82 \\
-FP & 0.911 & 0.098 & 55.97 & 45.34 \\
$\times$Net-I-22k & 0.912 & 0.096 & 56.57 & 45.80 \\
$\times$Net-22k & 0.916 & 0.095 & 56.82 & 45.91 \\
\midrule
\midrule
$\times$Net-I-0 & 0.899 & 0.104 & 52.72 & 42.67 \\
$\times$Net-I-1k & 0.908 & 0.100 & 55.50 & 44.83 \\
$\times$Net-I-22k & 0.912 & 0.096 & 56.57 & 45.80 \\
\midrule
\end{tabular}
}
\caption{
Ablation study and pre-training comparisons on NYU-Depth-V2 MDE and ADE20k sematic segmentation.
The first group is the ablation study result.
The second group involves comparisons among various decoders based on their pre-training volumes. Specifically, ``0'' indicates no pre-training, while ``1k'' and ``22k'' refer to pre-training on ImageNet-1k and ImageNet-22k, respectively.
}
\label{tab:ab}
\end{table}

\subsection{Ablation study}
In this subsection, we progressively modify an FPN into $\times$Net, with the results presented in the first group of Table~\ref{tab:ab}.
As shown in the first row of Table~\ref{tab:ab}, the ConvNeXt-T structure is not well-suited for dense prediction decoding. 
Reversing the decoding structure (from high resolution to low resolution) results in a slight improvement in MDE but a decrease in performance in semantic segmentation.
Once the decoding structure is reversed, pre-trained parameters can be loaded into the decoder. 
Using ImageNet-22k pre-training in the decoder significantly enhances model performance (reducing Abs Rel by 4.81\% and improving mIoU by 4.78\%).
Adding PFP and removing FP yield slight improvement. 
Mixing feature maps into image processing aids the transfer of pre-learned representations to the decoder: the mixing method of $\times$Net-I reduces Abs Rel by by 2.04\% and improve mIoU by 1.01\%, while the mixing method of $\times$Net reduces Abs Rel by 3.06\%  and improve mIoU by 1.26\%.

In summary,
it is evident that the pre-trained decoder stands as a pivotal element in the architectural design of $\times$Net.
Consequently, the next subsection will delve into an analysis of the mechanisms through which the pre-trained decoder enhances the efficacy of $\times$Net.

\subsection{Why decoder require pre-training?}
\label{sec:4.2}
In this subsection, we utilize $\times$Net-I to investigate the role of pre-trained decoders on network performance.
We conduct both quantitative and qualitative analyses.
We compare same decoder with different initializations: randomly initialized, pre-trained on ImageNet-1k classification, and pre-trained on ImageNet-22k classification.

\textbf{Quantitative analyse.}
As shown in the second group of Table~\ref{tab:ab}, compared to the non-pre-trained decoder, the ImageNet-1k pre-trained decoder achieves a 4.84\% reduction in Abs Rel, along with improvements of 5.27\% in mAcc and 5.06\% in mIoU. In contrast, the ImageNet-22k pre-trained decoder demonstrates significant enhancements, with a 7.69\% reduction in Abs Rel and improvements of 7.30\% in mAcc and 7.34\% in mIoU.

\textbf{Quantitative analyses.}
For quantitative analyses,
we visualize the mixed features and the final predictions.
For $\times$Net-I, due to its mixed feature being a three-channel feature map integrated into the RGB image, it inherently provides a visualization advantage.

The qualitative comparisons are shown in Figure~\ref{fig:nyu&seg}.
There is an interesting observation: despite utilizing the same encoder, differences in the pre-training of the decoder resulted in significant variations in the mixed features.
It is evident that the features of the decoder without pre-training are mostly concentrated in the same channels, whereas the pre-trained decoder allows for the observation of semantic information in the mixed feature. 
As explained in Subsection~\ref{sec:3.2}, $\times$Net-I aligns the optimization of pre-trained decoder closely with image processing.
As a result, the pre-learned representations in the decoder are effectively transferred into the decoding process.
For instance, the sofa in the third row of the MDE as shown in Figure~\ref{fig:nyu&seg} is well-separated into channels distinct from surrounding objects.
This separation is more pronounced with the ImageNet-22k pre-trained decoder compared to the ImageNet-1k pre-trained decoder, indicating higher confidence with ImageNet-22k pre-training.
Given that MDE supervises the depth prediction and the model cannot gain semantic information from MDE training.
Such clustering of semantic information can only be derived from the knowledge gained during pre-training.

As illustrated by the prediction comparisons in Figure~\ref{fig:nyu&seg}, the inclusion of semantic information in the mixed feature contributes to improvements beyond mere accuracy.
In MDE predictions, decoders without pre-training produce coarser outputs, where noticeable jagged edges can be observed upon magnification. In contrast, predictions from pre-trained decoders exhibit sharper details.
For semantic segmentation, there is a more coherent output with pre-trained decoders, whereas decoders without pre-training tend to predict different categories within the same object.
Pre-trained decoders significantly mitigate this issue. The semantic information embedded in the mixed feature enables the decoding process to effectively leverage pre-learned representations for improved prediction.

It can be concluded that pre-trained knowledge is effectively transferable to the decoder, endowing the network's mixed features with semantic information. This semantic enrichment can be leveraged in dense prediction tasks to provide sharper details and more coherent predictions.

\section{Dense prediction performances}
In this section, we compare $\times$Net with other advanced methods. 
To thoroughly evaluate the advantages of the pre-trained decoder, \textbf{we do not incorporate task-specific designs in $\times$Net}.
Instead, we employed a simple linear probe for prediction.
To ensure a fair comparison, we use the same encoder as the main competing methods, including identical model architecture and encoder pre-training.
Training was conducted using only standard data augmentation without introducing additional strategies. Moreover, we strictly controlled the computational cost of $\times$Net using FPS as a metric, ensuring that performance improvements are entirely due to the advantages of the pre-trained decoder.

\begin{table}
\resizebox{\columnwidth}{!}{
\begin{tabular}{l|l|ccc}
\midrule
Method & 
Backbone &
$\delta < 1.25 \uparrow$ &
Abs Rel$\downarrow$ & log10$\downarrow$ \\
\midrule
\midrule
DepthFormer \cite{li2023depthformer} & Swin-L~\cite{liu2021swin} & 0.921 &  0.096 & 0.041 \\
NeW CRFs~\cite{yuan2022newcrfs} & Swin-L~\cite{liu2021swin} & 0.922 & 0.095 & 0.041 \\
BinsFormer~\cite{li2024binsformer} & Swin-L~\cite{liu2021swin} & 0.925 & 0.094 & 0.040 \\
PixelFormer \cite{agarwal2023attention} & Swin-L~\cite{liu2021swin} & 0.929 & 0.090 & 0.039 \\
MG~\cite{liu2023single} & Swin-L~\cite{liu2021swin} & 0.933 & 0.087 & - \\
NDDepth~\cite{shao2023nddepth} & Swin-L~\cite{liu2021swin} & \textbf{0.936} & 0.087 & 0.038 \\
$\times$ConvNeXtV2-T (ours) & Swin-L~\cite{liu2021swin} & \textbf{0.936} & \textbf{0.086} & \textbf{0.037} \\
\midrule
\midrule
Depth Anything~\cite{yang2024depth} & DINOv2-L~\cite{oquab2023dinov2} & 0.984 & 0.056 & 0.024 \\
Metric3Dv2~\cite{hu2024metric3d} & DINOv2-L~\cite{oquab2023dinov2} & 0.989 & 0.047 & \textbf{0.020} \\
$\times$ConvNeXtV2-B (ours) & DINOv2-L~\cite{oquab2023dinov2} & \textbf{0.990} & \textbf{0.045} & \textbf{0.020} \\
\midrule
\end{tabular}
}
\caption{
MDE results on the NYU-Depth-V2 dataset.
}
\label{tab:nyu2}
\end{table}

\begin{table}
\resizebox{\columnwidth}{!}{
\begin{tabular}{l|l|ccc}
\midrule
Method & 
Backbone &
$\delta < 1.25 \uparrow$ &
Abs Rel$\downarrow$ & 
RMS\_log$\downarrow$ \\
\midrule
\midrule
NeW CRFs~\cite{yuan2022newcrfs} & Swin-L~\cite{liu2021swin} & 0.974 & 0.052 & 0.079 \\
BinsFormer~\cite{li2024binsformer} & Swin-L~\cite{liu2021swin} & 0.974 & 0.052 & 0.079 \\
DepthFormer \cite{li2023depthformer} & Swin-L~\cite{liu2021swin} & 0.975 &  0.052 & 0.079 \\
PixelFormer~\cite{agarwal2023attention} & Swin-L~\cite{liu2021swin} & 0.976 & 0.051 & 0.077 \\
MG\cite{liu2023single} & Swin-L~\cite{liu2021swin} & 0.976 & 0.050 & \textbf{0.075} \\
NDDepth~\cite{shao2023nddepth} & Swin-L~\cite{liu2021swin} & \textbf{0.978} & 0.050 & \textbf{0.075} \\
$\times$ConvNeXt-B (ours) & Swin-L~\cite{liu2021swin} & \textbf{0.978} & \textbf{0.048} & \textbf{0.075} \\
\midrule
\midrule
Depth Anything*~\cite{yang2024depth} & DINOv2-L~\cite{oquab2023dinov2} & 0.982 & 0.046 & 0.069 \\
Metric3Dv2*~\cite{hu2024metric3d} & DINOv2-L~\cite{oquab2023dinov2} & 0.985 & 0.044 & 0.060 \\
$\times$ConvNeXtV2-B (ours) & DINOv2-L~\cite{oquab2023dinov2} & \textbf{0.988} & \textbf{0.037} & \textbf{0.056} \\
\midrule
\end{tabular}
}
\caption{
MDE results on the KITTI dataset.
}
\label{tab:kitti2}
\end{table}

\begin{table}
\resizebox{\columnwidth}{!}{
\begin{tabular}{l|l|ccc}
\midrule
Method & 
Backbone &
$\delta < 1.25 \uparrow$ &
Abs Rel$\downarrow$ & log10$\downarrow$ \\
\midrule
\midrule
BinsFormer~\cite{li2024binsformer} & Swin-L~\cite{liu2021swin} & 0.805 & 0.143 & 0.061  \\
PixelFormer \cite{agarwal2023attention} & Swin-L~\cite{liu2021swin} & 0.802 & 0.144 & 0.062 \\
DepthFormer \cite{li2023depthformer} & Swin-L~\cite{liu2021swin} & 0.815 &  0.137 & 0.060 \\
NDDepth~\cite{shao2023nddepth} & Swin-L~\cite{liu2021swin} &    0.820 & 0.137 & 0.060 \\
$\times$ConvNeXtV2-T (ours) & Swin-L~\cite{liu2021swin} & \textbf{0.824} & \textbf{0.133} & \textbf{0.059} \\
\midrule
\end{tabular}
}
\caption{
Zero-shot performances on the SUN RGB-D dataset.
}
\label{tab:sun}
\end{table}

\subsection{Quantitative comparisons}
\textbf{NYU-Depth-V2 and KITTI results.} 
Table~\ref{tab:nyu2} and Table~\ref{tab:kitti2} compare the monocular depth estimation results of $\times$Net with methods specifically designed for monocular depth estimation, and $\times$Net shows state-of-the-art performance.
With the same Swin-L backbone, it surpasses NDDepth on both the NYU-Depth-V2 dataset (0.086 Abs Rel vs. 0.087 Abs Rel) and the KITTI dataset (0.048 Abs Rel vs. 0.050 Abs Rel).
Unlike NDDepth, which requires using camera intrinsics to compute Normal-Distance for depth prediction, our method simply employs a pre-trained ConvNeXtV2-T~\cite{woo2023convnext} or ConvNeXt-B~\cite{liu2022convnet} for decoding.
For the Metric3Dv2~\cite{hu2024metric3d} trained on large-scale datasets, we achieve significant improvements over Metric3Dv2's fine-tuning results on the KITTI dataset (0.037 Abs Rel vs. 0.044 Abs Rel) using Metric3Dv2 pre-trained DINOv2 $\times$ ConvNeXtV2-B~\cite{woo2023convnext}.

\textbf{Zero-shot results of SUN RGB-D and online evaluation on KITTI server.}
Following previous works~\cite{li2024binsformer, agarwal2023attention, li2023depthformer, shao2023nddepth}, we evaluate the NYU-Depth-V2 trained model on the SUN RGB-D dataset. The results in Table~\ref{tab:sun} show that our pre-trained ConvNeXtV2-T decoder achieved the best generalization performance, surpassing previous methods across all three metrics.
Additionally, we uploaded the KITTI benchmark training results to the KITTI server for validation.
At the time of submission, our model \textbf{ranks 1st} among all submissions.
As shown in Table~\ref{tab:kitti2}, our method significantly outperforms the previous state-of-the-art method~\cite{piccinelli2024unidepth} across all four metrics.

\begin{table}
\resizebox{\columnwidth}{!}{
\begin{tabular}{l|cccc}
\midrule
Method & \small{SILog}$\downarrow$ & \small{sqRel}$\downarrow$ & \small{absRel}$\downarrow$ & \small{iRMSE}$\downarrow$ \\
\midrule
\midrule
NeW CRFs \cite{yuan2022newcrfs} & 10.39 & 1.83 & 8.37 & 11.03 \\
PixelFormer \cite{li2024binsformer} & 10.28 & 1.82 & 8.16 & 10.84 \\
BinsFormer \cite{li2024binsformer} & 10.14 & 1.69 & 8.23 & 10.90 \\
DepthFormer~\cite{li2024binsformer} & 10.69 & 1.84 & 8.68 & 11.39 \\
MG~\cite{liu2023single} & 9.93 & 1.68 & 7.99 & 10.63 \\
NDDepth~\cite{shao2023nddepth} & 9.62 & 1.59 & 7.75 & 10.62 \\
UniDepth~\cite{piccinelli2024unidepth} & 8.13 & 1.09 & 6.54 & 8.24 \\
D2L$\times$C2B & \textbf{7.51} & \textbf{0.93} & \textbf{6.14} & \textbf{7.62}  \\
\midrule
\end{tabular}
}
\caption{
MDE performances on the online KITTI evaluation server.
``D2L$\times$C2B'' denotes DINOv2-L encoder $\times$ ConvNeXtV2-B decoder.
}
\label{tab:server}
\end{table}

\begin{figure*}
\centering
\includegraphics[width=\textwidth]{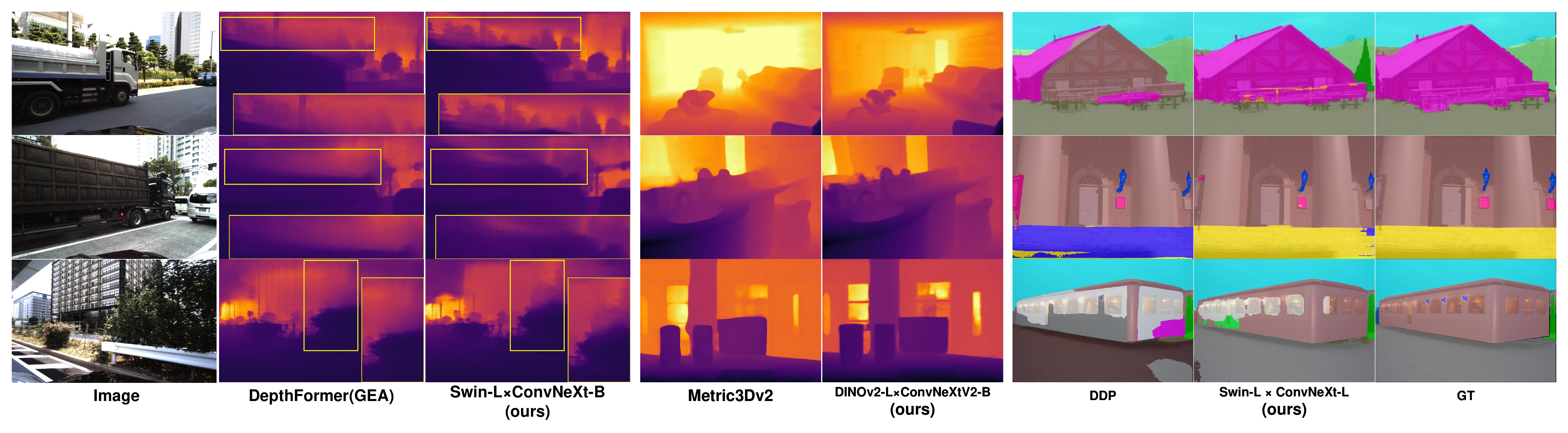}
\caption{
Qualitative comparisons on DDAD MDE (left) and NYU-Depth-V2 MDE (middle) and ADE20k semantic segmentation (right).
}
\label{fig:quality}
\end{figure*}

\begin{table}
\resizebox{\columnwidth}{!}{
\begin{tabular}{l|l|ccc}
\midrule
Method & 
Backbone &
Abs Rel$\downarrow$ & 
Sq Rel$\downarrow$ & 
RMS\_log$\downarrow$ \\
\midrule
\midrule
DepthFormer~\cite{li2023depthformer} & Swin-L~\cite{liu2021swin} & 0.152 &  2.230 & 0.246 \\
PixelFormer~\cite{agarwal2023attention} & Swin-L~\cite{liu2021swin} & 0.151 & 2.140 & 0.242 \\
BinsFormer~\cite{li2024binsformer} & Swin-L~\cite{liu2021swin} &  0.149 & 2.142 & 0.244 \\
DepthFormer(GEV)$\dagger$~\cite{yang2023gedepth} & Swin-L~\cite{liu2021swin} & 0.149 & 2.121 & 0.240 \\
DepthFormer(GEA)$\dagger$~\cite{yang2023gedepth} & Swin-L~\cite{liu2021swin} & 0.145 & 2.119 & 0.237 \\
$\times$ConvNeXt-B (ours) & Swin-L~\cite{liu2021swin} & \textbf{0.144} & \textbf{2.116} & \textbf{0.235} \\
\midrule
\end{tabular}
}
\caption{
MDE performances on the DDAD dataset.
Methods using extra camera parameter inputs are marked with ``$\dagger$''.
}
\label{tab:ddad}
\end{table}

\textbf{DDAD results.} 
Table~\ref{tab:ddad} presents the results on the DDAD dataset.
$\times$Net significantly outperforms many representative models~\cite{li2023depthformer, agarwal2023attention, li2024binsformer}.
Furthermore, even when DepthFormer incorporates ground embedding~\cite{yang2023gedepth}, which requires extra camera parameter inputs, our model achieves better results with a simple combination of Swin-L $\times$ ConvNeXt-B.

\textbf{ADE20K results.}
Table~\ref{tab:ade2} presents the comparison results on the ADE20K dataset, where our combination of Swin-L$\times$ConvNeXt-L achieved the best results.
Using only a 512$\times$512 crop size, we surpassed UperNet trained with a 640$\times$640 crop size using Swin-L, ConvNeXt-L, and ConvNeXt-XL backbones.
Additionally, this simple combination also outperformed the 3-step diffusion model, DDP~\cite{ji2023ddp}.

\textbf{Cityscapes results.}
As shown in Table~\ref{tab:city}, $\times$Net surpasses various specialized semantic segmentation methods on the Cityscapes dataset.
By using ConvNeXt-L for decoding, it exceeds the performance of Mask2Former~\cite{cheng2022masked} (83.63 mIoU vs. 83.30 mIoU), even when Mask2Former employs a dual decoder structure with pixel and Transformer decoders and is trained with additional loss functions, all while using the same backbone.

\begin{table}
\resizebox{\columnwidth}{!}{
\begin{tabular}{l|lc|c}
\midrule
Method & 
Backbone &
crop size &
mIoU $\uparrow$ \\
\midrule
\midrule
SegFormer~\cite{xie2021segformer} & MiT-B5~\cite{xie2021segformer} & 512$^2$ & 51.0  \\
Swin-UperNet~\cite{liu2021swin, xiao2018unified} & Swin-L~\cite{liu2021swin} & 640$^2$ & 52.1 \\
ConvNeXt-UperNet~\cite{liu2022convnet, xiao2018unified} & ConvNeXt-L~\cite{liu2022convnet} & 640$^2$ & 52.1 \\
ConvNeXt-UperNet~\cite{liu2022convnet, xiao2018unified} & ConvNeXt-XL~\cite{liu2022convnet} & 640$^2$ & 53.6\\
MaskFormer~\cite{cheng2021per} & Swin-L~\cite{liu2021swin} & 640$^2$ & 54.1 \\
DDP (step 3)~\cite{ji2023ddp} & Swin-L~\cite{liu2021swin} & 512$^2$ & 53.2 \\
$\times$ConvNeXt-L (ours) & Swin-L~\cite{liu2021swin}  & 512$^2$ & \textbf{54.3} \\
\midrule
\end{tabular}
}
\caption{
Semantic segmentation performances on the ADE20k dataset.
}
\label{tab:ade2}
\end{table}


\begin{table}
\resizebox{\columnwidth}{!}{
\begin{tabular}{l|l|cc}
\midrule
Method & 
Backbone &
mIoU (s.s.) $\uparrow$ & 
mIoU (m.s.)$\uparrow$ \\
\midrule
\midrule
Segmenter~\cite{strudel2021segmenter} & ViT-L~\cite{dosovitskiy2020vit} & 79.10 &  81.30 \\
SETR-PUP~\cite{zheng2021rethinking} & ViT-L~\cite{dosovitskiy2020vit} & 79.34 &  82.15 \\
StructToken~\cite{lin2023structtoken} & ViT-L~\cite{dosovitskiy2020vit} & 80.05 &  82.07 \\
DDP (step 3)~\cite{ji2023ddp} & ConvNeXt-L~\cite{liu2022convnet} & 83.21 & 83.92 \\
DiversePatch~\cite{gong2021vision} & Swin-L~\cite{liu2021swin} & 82.70 &  83.60 \\
Mask2Former~\cite{cheng2022masked} & Swin-L~\cite{liu2021swin} & 83.30 &  84.30 \\
$\times$ConvNeXt-L (ours) & Swin-L~\cite{liu2021swin} & \textbf{83.63} &  \textbf{84.32} \\
\midrule
\end{tabular}
}
\caption{
Semantic segmentation performances on the Cityscapes dataset.
``s.s.'' and ``m.s.'' denote single-scale and multi-scale inference results, respectively.
}
\label{tab:city}
\end{table}

\begin{table}
\begin{center}
\resizebox{\columnwidth}{!}{
\begin{tabular}{l|ccc}
\midrule
Model & Backbone & Dataset & FPS (image/s) \\
\midrule
\midrule
BinsFormer~\cite{li2024binsformer} & Swin-L~\cite{liu2021swin} & NYU & 2.20$^\dagger$ \\
NDDepth~\cite{shao2023nddepth} & Swin-L~\cite{liu2021swin} & NYU & 2.23$^\dagger$ \\
$\times$ConvNeXtV2-T & Swin-L~\cite{liu2021swin} & NYU & \textbf{5.21$^\dagger$} \\
\midrule
\midrule
Depth Anything~\cite{yang2024depth} & DINOv2-L~\cite{oquab2023dinov2} & NYU & 1.85$^\dagger$ \\
Metric3Dv2~\cite{hu2024metric3d} & DINOv2-L~\cite{oquab2023dinov2} & NYU & 0.89$^\dagger$ \\
$\times$ConvNeXtV2-B & DINOv2-L~\cite{oquab2023dinov2} & NYU & \textbf{2.00$^\dagger$} \\
\midrule
\midrule
BinsFormer~\cite{li2024binsformer} & Swin-L~\cite{liu2021swin} & KITTI & 1.70$^\dagger$ \\
NDDepth~\cite{shao2023nddepth} & Swin-L~\cite{liu2021swin} & KITTI & 1.82$^\dagger$ \\
$\times$ConvNeXt-B & Swin-L~\cite{liu2021swin} & KITTI & \textbf{3.32$^\dagger$} \\
\midrule
\midrule
Depth Anything~\cite{yang2024depth} & DINOv2-L~\cite{oquab2023dinov2} & KITTI & 0.46$^\dagger$ \\
Metric3Dv2~\cite{hu2024metric3d} & DINOv2-L~\cite{oquab2023dinov2} & KITTI & 0.63$^\dagger$ \\
$\times$ConvNeXtV2-B & DINOv2-L~\cite{oquab2023dinov2} & KITTI & \textbf{1.33$^\dagger$} \\
\midrule
\midrule
ConvNeXt-UperNet~\cite{liu2022convnet,xiao2018unified} & ConvNeXt-XL~\cite{liu2022convnet} & ADE20K & 13.97$^\ddagger$ \\
DDP~\cite{ji2023ddp} & Swin-L~\cite{liu2021swin} & ADE20K & 15.07$^\ddagger$ \\
$\times$ConvNeXt-L & Swin-L~\cite{liu2021swin} & ADE20K & \textbf{15.74$^\ddagger$} \\
\midrule
\midrule
DDP~\cite{ji2023ddp} & Swin-L~\cite{liu2021swin} & CityScapes & 4.15$^\ddagger$ \\
$\times$ConvNeXt-L & Swin-L~\cite{liu2021swin} & CityScapes & \textbf{5.95$^\ddagger$} \\
\midrule
\end{tabular}
}
\end{center}
\caption{
Efficiency comparisons.
$^\dagger$ denotes FPS is evaluated on a RTX 2080 Super GPU,
while $^\ddagger$ denotes FPS is evaluated on a Tesla A100 GPU GPU,
.
}
\label{tab:mdefps}
\end{table}

\subsection{Qualitative analysis and efficiency comparison}
As illustrated in Figure 5, and as mentioned in Subsection~\ref{sec:4.2}, pre-trained decoder predictions exhibit more robust semantic information compared to other methods, resulting in sharper and more coherent predictions.
For instance, in the DDAD MDE predictions, Swin-L $\times$ ConvNeXt-B demonstrates sharper detail than DepthFormer (GEA), maintaining precise contours for both large and small objects. In the DDAD MDE predictions, Metric3Dv2 is noted to cause some object deformations, whereas Swin-L $\times$ ConvNeXt-L preserves accurate outlines. In the ADE20k semantic segmentation comparisons, DDP tends to show inconsistent errors in the central areas of large objects, yet Swin-L $\times$ ConvNeXt-L maintains consistent predictions.

To ensure a fair comparison, we avoided performance improvements that might arise from using computationally intensive decoders in the decoder section. Table~\ref{tab:mdefps} compares the computational efficiency of the primary competing methods, demonstrating that $\times$Net consistently maintains the highest efficiency across all comparisons.

\section{Conclusion}
This paper is the first to introduce the use of a pre-trained decoder in an encoder-decoder structure, termed $\times$Net.
It achieves efficient dense prediction performance through the ``pre-trained encoder $\times$ pre-trained decoder'' collaboration.
The pre-trained decoder enriches the network's intermediate features with semantic information, which improves the accuracy and the quality of details of dense predictions.


{
    \small
    \bibliographystyle{ieeenat_fullname}
    \bibliography{main}
}

\end{document}